\documentclass{article}

\usepackage{microtype}
\usepackage{graphicx}
\usepackage{subfigure}
\usepackage{booktabs} 

\usepackage{epstopdf}
\usepackage{amsfonts}
\usepackage{amssymb}
\usepackage{amsthm}
\usepackage{amsmath}
\usepackage{url}
\usepackage{multirow}
\usepackage{array}
\usepackage[accepted]{icml2018}

\usepackage{enumitem}
\setlist{leftmargin=3.5mm}

\def\etal{{\em et al.\ }}

\newcommand{\NM}[2]{\| #1 \|_{#2}}
\newcommand{\NML}[2]{\left\| #1 \right\|_{#2}}
\newcommand{\R}{\mathbb{R}}

\newcommand{\iFFT}[1]{\mathcal{F}^{-1} ( #1 )} 
\newcommand{\FFT}[1]{\mathcal{F}( #1 )} 
\newcommand{\CC}[1]{\mathcal{C}( #1 )}
\newcommand{\Px}[2]{\text{prox}_{#1}( #2 ) }

\newtheorem{theorem}{Theorem}

\newtheorem{proposition}[theorem]{Proposition}

\usepackage{array}
\newcolumntype{L}[1]{>{\raggedright\let\newline\\\arraybackslash\hspace{0pt}}m{#1}}
\newcolumntype{C}[1]{>{\centering\let\newline  \\\arraybackslash\hspace{0pt}}m{#1}}
\newcolumntype{R}[1]{>{\raggedleft\let\newline \\\arraybackslash\hspace{0pt}}m{#1}}
\icmltitlerunning{Online Convolutional Sparse Coding with Sample-Dependent Dictionary}
\begin{document}

\twocolumn[
\icmltitle{Online Convolutional Sparse Coding with Sample-Dependent Dictionary}



\icmlsetsymbol{equal}{*}

\begin{icmlauthorlist}
\icmlauthor{Yaqing Wang}{hkust}
\icmlauthor{Quanming Yao}{hkust,4pa}
\icmlauthor{James T. Kwok}{hkust}
\icmlauthor{Lionel M. Ni}{mac}
\end{icmlauthorlist}

\icmlaffiliation{hkust}{Department of Computer Science and Engineering, 
	Hong Kong University of Science and Technology University, Hong Kong}
\icmlaffiliation{mac}{Department of Computer and Information Science,
	University of Macau, Macau}
\icmlaffiliation{4pa}{4Paradigm Inc, Beijing, China.}

\icmlcorrespondingauthor{jamesk@cse.ust.hk}{James T. Kwok}

\icmlkeywords{Online Learning, Convolutional Sparse Coding, Sample-Dependent Dictionary Learning.}

\vskip 0.3in
]



\printAffiliationsAndNotice{}  

\begin{abstract}
Convolutional sparse coding (CSC) has been popularly used
for the
learning of shift-invariant dictionaries
in image and signal processing.
However, existing methods have limited scalability.
In this paper, instead of convolving with a dictionary shared by all samples, we 
propose the use of 
a sample-dependent dictionary in which
each filter is a linear combination of a small set of 
base filters learned from data.
This added flexibility 
allows
a large number of sample-dependent patterns 
to be captured,
which is especially useful in the handling of large 
or high-dimensional 
data sets.
Computationally, the resultant model can be efficiently learned by online learning. 
Extensive experimental results on a number of data sets
show that the proposed method 
outperforms existing CSC algorithms with significantly reduced time and space
complexities.
\end{abstract}


\section{Introduction}

Convolutional sparse coding (CSC) 
\cite{zeiler2010deconvolutional} 
has been successfully used 
in image processing 
\cite{gu2015convolutional,heide2015fast}
signal processing \cite{cogliati2016context},
and 
biomedical applications
\cite{pachitariu2013extracting,andilla2014sparse,chang2017unsupervised,jas2017learning,peter2017sparse}.
It is closely related to 
sparse coding \cite{aharon2006rm}, but
CSC  is advantageous in that its
shift-invariant 
dictionary 
can capture shifted local patterns common in signals and images.
Each data sample is then represented by the sum of a set of filters from the dictionary
convolved with the corresponding
codes.


%

Traditional CSC algorithms operate in the batch mode
\cite{kavukcuoglu2010learning,zeiler2010deconvolutional,
bristow2013fast,heide2015fast,sorel2016fast,wohlberg2016efficient,papyan2017convolutional}, which
take
$O(NK^2P+NKP\log P)$ time
and $O(NKP)$ space
(where $N$ is the number of samples,
$K$ is the number of filters, and $P$ is the dimensionality.
Recently, a number of online CSC algorithms 
have been 
proposed
for better scalability
\cite{degraux2017online,liu2017online,wang2018online}.
As data samples arrive, 
relevant information is compressed into small history statistics,
and 
the model is incrementally updated.
In particular, the state-of-the-art OCSC algorithm 
\cite{wang2018online}
has the
much smaller 
time and
space complexities
of $O(K^2P+KP\log P)$ 
and 
$O(K^2P)$,  respectively.

However,
the 
complexities of OCSC still depend quadratically on $K$,
and cannot be used with a large number of filters. The number of local
patterns that can be captured is thus limited, and may lead to inferior
performance especially on higher-dimensional data sets.  Besides, the use of more filters
also leads to a larger number
of expensive convolution operations. 
\citeauthor{rigamonti2013learning} 
\yrcite{rigamonti2013learning} and 
\citeauthor{sironi2015learning} 
\yrcite{sironi2015learning} proposed to post-process the learned filters by
approximating them with separable filters,
making the convolutions less expensive. However, as
learning and post-processing are then two independent procedures, the resultant
filters may not be optimal. Moreover, these separate filters cannot be updated online
with the arrival of new samples.

Another direction to scale up CSC is via
distributed computation \cite{bertsekas1997parallel}. 
By distributing the data and workload onto multiple machines,
the recent
consensus CSC algorithm
\cite{choudhury2017consensus} 
can handle large, higher-dimensional data sets such as videos, multispectral images and light field images.
However, the 
heavy computational demands of the CSC problem are only shared over the computing platform, but not 
fundamentally
reduced.


In this paper, 
we propose to approximate the possibly large number of filters by a sample-dependent combination of a small set of 
base filters learned from the data.
While the standard CSC
dictionary is shared by all samples,
we propose 
each sample to have its own ``personal" dictionary 
to compensate for the reduced flexibility of using these base filters. 
In this way,  
the representation power can remain the same but with
a reduced
number of parameters.
Computationally, this structure also allows 
efficient online learning algorithms to be developed. Specifically,
the base filter can be updated by 
the alternating direction method of multipliers (ADMM)
\cite{boyd2011distributed},
while
the codes and combination weights can be learned
by 
accelerated proximal algorithms
\cite{yao2017efficient}.
Extensive experimental results on a variety of data sets
show that the proposed algorithm 
is more efficient in both time and space,
and outperforms existing batch, online and distributed CSC algorithms.






The rest of the paper is organized as follows. Section~\ref{sec:related}
briefly reviews convolutional sparse coding. 
Section~\ref{sec:method} describes the proposed algorithm.
Experimental results
are presented in Section~\ref{sec:expt}, and
the last section gives some
concluding remarks.

%


\noindent
{\bf Notations}: 
For a vector $a$,
its $i$th element is $a(i)$,
$\ell_1$-norm is $\|a\|_1 = \sum_i |a(i)|$ 
and 
$\ell_2$-norm is $\|a\|_2 =
\sqrt{\sum_i a^2(i)}$.
The convolution of two vectors $a$ and $b$ is 
denoted 
$a*b$.
For a matrix $A$,
$A^\star$ is its complex conjugate, and
$A^\dagger=
({A^\top})^\star$
its conjugate transpose.
The Hadamard product 
of two matrices $A$ and $B$
is
$( A\odot B )(i,j) = A(i,j) B(i,j)$.
The identity matrix is denoted $I$. 
$\FFT{x}$
is 
the Fourier transform
that maps $x$ from the spatial domain to the frequency
domain, 
while 
$\iFFT{x}$ is the inverse operator which maps $\FFT{x}$ back to $x$. 


\section{Review: Convolutional Sparse Coding}
\label{sec:related}

Given samples $\{x_1,\dots,x_N\}$ in $\mathbb{R}^{P}$,
CSC learns a shift-invariant dictionary 
$D\in\R^{M\times K}$,
with the columns $\{D(:,k)\}$ representing the $K$ filters.
Each sample $x_i$ is encoded as $Z_i \in\R^{P\times K}$, with the $k$th column being the code convolved with 
filter 
$D(:,k)$. 
The dictionary and codes are learned together by solving
the optimization problem:
\begin{equation}
\label{eq:csc}
\!
\min_{D \in \mathcal{D}, \{Z_i\}}
\!\!
\frac{1}{N}
\!\sum_{i=1}^{N}\!
\frac{1}{2}
\NML{ x_i
\!-\!\!
\sum_{k=1}^{K} \! D(:,k)
\!*\!
Z_i(:,k) 
}{2}^2
\!\!\! +  
\beta \NM{Z_i}{1},\!
\end{equation}
where the first term measures the signal reconstruction error, $\mathcal{D} = \{ D : \NM{ D(:,k) }{2} \le 1, \forall k = 1,\dots, K \}$ ensures
that the filters are normalized, and $\beta
>0$ is a regularization parameter controlling the sparsity of
$Z_i$'s.

Convolution 
in \eqref{eq:csc}
is 
performed in the spatial domain. This takes $O(KPM)$ time, and is
expensive.
In contrast,
recent CSC methods 
perform convolution 
in the frequency domain, which
takes $O(KP\log P)$ time
\cite{mallat1999wavelet} and is faster for typical choices of 
$M$ 
and $P$.
Let $\tilde{x}_i \equiv \FFT{x_i}$,
$\tilde{D}(:,k) \equiv
\FFT{D(:,k)}$,
and
$\tilde{Z}_i(:,k) \equiv \FFT{Z_i(:,k)}$ be the Fourier-transformed counterparts of
$x_i, D(:,k)$ and $Z_i(:,k)$. 
The codes and filters 
are 
updated
in an alternating manner
by block coordinate descent, as:

\noindent
\textbf{Update Codes:}
Given 
$\tilde{D}$, each
$\tilde{Z}_i$ is independently obtained as
\begin{align}
\!\!\!\!
\min_{\tilde{Z}_i,{U}_i}
&~~
\frac{1}{2P} 
\NML{\tilde{x}_i
	\!-\!
	\sum_{k=1}^{K}\tilde{D}(:,k)
	\odot
	\tilde{Z}_i(:,k)}{2}^2
+\beta \NM{U_i}{1}
\label{eq:csc_code} \\
\text{s.t.} &~~ 
\FFT{U_i(:,k)} = \tilde{Z}_i(:,k)
, 
\; k =1,\dots,K,
\notag
\end{align} 
where $U_i$ is an auxiliary variable.

\noindent
\textbf{Update Dictionary:}
$\tilde{D}$ is 
updated by solving
\begin{eqnarray}
&\min_{\tilde{D},V}
& \!\!\!\!\!\!
\frac{1}{2NP} \sum_{i=1}^{N} 
\NML{\tilde{x}_i
	-
	\sum_{k=1}^{K}\tilde{D}(:,k)
	\odot
	\tilde{Z}_i(:,k)}{2}^2
\label{eq:csc_dic}
\\
&\text{s.t.}
&\!\!\!\!\!\!
\FFT{V(:,k)} = \tilde{D}(:,k), 
\;k =1,\dots,K, 
\notag
\\
&&\!\!\!\!\!\! \NM{\CC{V(:,k)}}{2}^2 \le 1, 
\; k =1,\dots,K,
\nonumber
\end{eqnarray}
where $V$ is an auxiliary variable,
and
$\mathcal{C}(\cdot)$ 
crops the extra $P-M$ dimensions in
${{V}(:,k)}$.

Both 
(\ref{eq:csc_code}) and
(\ref{eq:csc_dic})
can be
solved by 
the alternating direction method of multipliers (ADMM) 
\cite{boyd2011distributed}.
Subsequently,
$\{\tilde{Z}_i\}$ and $\tilde{D}$
can be transformed back to the spatial domain as $Z_i(:,k) = \iFFT{\tilde{Z}_i(:,k)}$ and
$D(:,k) = \mathcal{C}(\iFFT{\tilde{D}(:,k))}$.
Note that 
while $Z_i$'s 
(in the spatial domain)
are sparse, the FFT-transformed 
$\tilde{Z}_i$'s
are not.

On inference,
given the learned dictionary $D$,
the testing sample $x_j$ is reconstructed as $\sum_{k=1}^{K}D(:,k)*Z_j(:,k)$,
where $Z_j$ is the obtained code.


\subsection{Post-Processing for Separable Filters}
\label{sec:post}

Filters obtained by CSC are non-separable and subsequent convolutions may be slow.
To speed this up, they can be post-processed and approximated by separable filters
\cite{rigamonti2013learning,sironi2015learning}.
Specifically, the learned $D$ is approximated by $SW$, where 
$S\in \R^{M\times R}$ contains $R$ rank-1 base filters
$\{S(:,1),\dots,
S(:,R)\}$, and
$W\in\R^{R\times K}$ contains the combination
weights.
However, this
often leads to 
performance
deterioration.



\subsection{Online CSC}
\label{sec:online}


An online CSC algorithm (OCSC)
is recently proposed in \cite{wang2018online}.
Given the 
Fourier-transformed 
sample $\tilde{x}_t$ and
dictionary 
$\tilde{D}_{t-1}$ from the last iteration,
the corresponding
$\{\tilde{Z}_t, U_t\}$ are obtained as in \eqref{eq:csc_code}.
The following Proposition 
updates
$\tilde{D}_t$ and $V_t$ by 
reformulating
\eqref{eq:csc_dic} for use with smaller history statistics.

\begin{proposition}[\cite{wang2018online}]
	\label{pr:reorder}
	$\tilde{D}_t, V_t$ can be obtained by solving the optimization problem:
	\begin{eqnarray}
	&\min_{\tilde{D},V}
	&
	\frac{1}{2P}
	\sum_{p=1}^{P}
	\tilde{D}(p,:)H_t(:,:,p) \tilde{D}^{\dagger}(:,p)
	\nonumber
	\\
	&&
	-2\tilde{D}(p,:)G_t(:,p)
	\label{eq:ocsc_dic}
	\\
	&\text{s.t.}
	&
	\FFT{V(:,k)} = \tilde{D}(:,k), 
	\; k =1,\dots,K, 
	\notag
	\\
	&& \NM{\CC{V(:,k)}}{2}^2 \le 1, 
	\; k =1,\dots,K,
	\nonumber
	\end{eqnarray}		
where $H_t(:,:,p)\!=\!\frac{1}{t}\sum_{i=1}^{t}\tilde{Z}_i^{\top}(:,p)\tilde{Z}^\star_i(p,:)
\!\in\!\mathbb{R}^{K\times K}$, 
	and
$G_t(:,p)\!=\!\frac{1}{t}\sum_{i=1}^{t}\tilde{x}^\star_i(p)\tilde{Z}_{i}^{\top}(:,p)
\!\in\!\mathbb{R}^{K}$.
\end{proposition}
Problem~(\ref{eq:ocsc_dic}) can be solved by ADMM.
The total space complexity is only $O(K^2P)$,
which is independent of $N$. Moreover, 
$H_t$ and $G_t$ can be updated incrementally.




Two other online CSC reformulations have also been proposed recently.
\citeauthor{degraux2017online} 
(\citeyear{degraux2017online})
performs convolution in the spatial domain, and is slow.
\citeauthor{liu2017online} 
(\citeyear{liu2017online})
performs convolution in the frequency domain, but 
requires 
expensive
huge sparse matrix operations.


\section{Online CSC with Sample-Dependent Dictionary}
\label{sec:method}

Though OCSC 
scales well with sample size $N$, 
its space complexity still depends on $K$ quadratically.
This limits the
number of filters that can be used and can impact performance.
Motivated by the ideas 
of separable filters 
in Section~\ref{sec:post},
we enable learning with more filters by approximating the $K$ filters with
$R$ base filters, where $R\ll K$.
In contrast to the 
separable filters,
which
are obtained
by post-processing and 
may not be optimal, 
we
propose to 
learn the dictionary 
directly during signal reconstruction.
Moreover, filters in the dictionary are 
combined from the base filters 
in a sample-dependent manner.


\subsection{Problem Formulation}
\label{sec:formulation}

Recall that 
each $x_i$ 
in (\ref{eq:csc})
is represented by $\sum_{k=1}^{K}D(:,k) * Z_i(:,k)$. Let $B\in\R^{M\times R}$, with columns $\{B(:,r)\}$ being the base filters.
We propose to represent $x_i$ as:
\begin{align}\label{eq:x_two_level}
x_i = \sum_{k=1}^{K}
\left( \sum_{r=1}^{R} W_i(r,k) {B}(:,r) \right) * Z_i(:,k), 
\end{align}
where
$W_i \in R^{R\times K}$ is the 
matrix for the 
filter combination weights.
In other words, each $D(:,k)$ in \eqref{eq:csc} is replaced by 
$\sum_{r=1}^{R} W_i(r,k) {B}(:,r)$,
or equivalently,
\begin{align}\label{eq:reduce_k}
D_i = B W_i, 
\end{align}
which is sample-dependent.
As will be seen,
this allows the
$W_i$'s to be 
learned
independently
(Section~\ref{sec:proalg}).
This also leads to 
more sample-dependent patterns 
being captured and thus better performance
(Section ~\ref{sec:expts_soa}).

Sample-dependent 
filters	
have been recently studied in convolutional neural networks 
(CNN)
\cite{jia2016dynamic}. 
Empirically, this
outperforms standard CNNs in 
one-shot learning \cite{bertinetto2016learning},
video prediction \cite{jia2016dynamic}
and image deblurring \cite{kang2017incorporating}.
\citeauthor{jia2016dynamic}
\yrcite{jia2016dynamic} 
uses a 
specially designed 
neural network
to learn the filters, and does not consider the CSC model.
In contrast, the sample-dependent filters here
are integrated into CSC.



The dictionary 
can also be adapted 
to individual samples 
by fine-tuning
\cite{donahue2014decaf}.
However, learning
the initial shared dictionary is still expensive when 
$K$ is 
large.
Besides,
as will be shown 
in Section~\ref{sec:expts_sample_dependent},
the proposed method 
outperforms fine-tuning
empirically.


\subsection{Learning}

Plugging 
(\ref{eq:reduce_k})
into the CSC formulation in \eqref{eq:csc}, we obtain
\begin{eqnarray} 
& \min_{B,\{W_i,Z_i\}} & \!\!\!
\frac{1}{N}
\sum_{i = 1}^N
f_i(B, W_i, Z_i) + \beta \NM{Z_i}{1}
\label{eq:orig}\\
& \text{s.t.} & \!\!\! BW_i\in\mathcal{D}, \; i=1,\dots,N, \label{eq:tmp1} \\
&& \!\!\! B\in\mathcal{B}, \label{eq:newb}
\end{eqnarray}
where 
\begin{align*} 
\!\!\!
f_i(B, W_i, Z_i)
\!\!\equiv\!\!
\frac{1}{2}
\!
\NML{
	x_i 
	\!\! -\! \! 
	\sum_{k=1}^{K}
	\!
	\! \left( 
	\! \sum_{r=1}^{R} \! W_i(r,k) {B}(:,r)
	\!\! \right) \!\! * \! Z_i(:,k)
}{2}^2,
\end{align*}
and $\mathcal{B} \equiv \{ B : \NM{ B(:,r) }{2} \le 1, \forall\; r = 1,\dots, R \}$.
As $B$ and $W_i$ are coupled together in \eqref{eq:tmp1},
this makes the optimization problem difficult.
The following Proposition decouples $B$ and $W_i$.
All the proofs are in the Appendix.

\begin{proposition} \label{pr:constraint}
	For $B\in\mathcal{B}$, we have
	$B W_i\in\mathcal{D}$ if
	(i) 
	$W_i \in \mathcal{W}_{\ell 1} \equiv \{ W : \NM{W_i(:,k)}{1} \le 1 , \;k = 1,\dots, K \}$,
	or 
	(ii)
	$W_i \in \mathcal{W}_{\ell 2} \equiv \{ W : \NM{W_i(:,k)}{2} \le 1 / \sqrt{R}, \;k = 1,\dots, K \}$.
\end{proposition}

To simplify notations,
we use $\mathcal{W}$ to denote $\mathcal{W}_{\ell 1}$ or $\mathcal{W}_{\ell 2}$.
By imposing either one of the above structures on $\{W_i\}$,
we have the following optimization problem:
\begin{eqnarray} 
& \min_{B,\{W_i,Z_i\}} &
\frac{1}{N}
\sum_{i = 1}^N
f_i(B, W_i, Z_i)
+ \beta \NM{ Z_i}{1}
\label{eq:cscsd_pre} 
\\
& \text{s.t.} & 
B\in\mathcal{B}, \;\text{ and } \;
W_i \in \mathcal{W}, i=1,\dots,N.
\notag
\end{eqnarray} 
On inference with sample $x_j$,
the corresponding $(W_j, Z_j)$ can be obtained by 
solving \eqref{eq:cscsd_pre}
with the 
learned $B$
fixed.


\subsection{Online Learning Algorithm for (\ref{eq:cscsd_pre}) }
\label{sec:proalg}

As in Section~\ref{sec:online}, 
we propose an online algorithm 
for better scalability.
At the $t$th iteration, 
consider
\begin{eqnarray} 
& \min_{B,\{W_i,Z_i\}} &
\frac{1}{t}
\sum_{i = 1}^t
f_i(B, W_i, Z_i)
+ \beta \NM{ Z_i}{1}
\label{eq:cscsd_pre_app} \\
& \text{s.t.} & 
B\in\mathcal{B}, \;\text{ and }\;
W_i \in \mathcal{W}, \; i=1,\dots,N. \nonumber 
\end{eqnarray} 

Let $\tilde{B}(:,r) \equiv \FFT{B(:,r)}$, where ${B}(:,r)\in\R^M$ is zero-padded to
be $P$-dimensional.
Note that 
the number of convolutions
can be 
reduced from $K$ to $R$
by rewriting
the summation above
as
$\sum_{r=1}^{R} B(:,r) * ( \sum_{k=1}^{K}W_i(r,k)Z_i(:,k) )$.
The following Proposition rewrites \eqref{eq:cscsd_pre_app} 	
and performs convolutions in the frequency domain.


\begin{proposition}\label{pr:cscsd}
	Problem~\eqref{eq:cscsd_pre_app} can be rewritten
	as
	\begin{eqnarray}
	\label{eq:cscsd}
	\!\!\!\!\!\!
	&\min_{\tilde{B}, \{W_i, Z_i\}}&
	\frac{1}{t}\sum_{i=1}^{t}
	\tilde{f}_i(\tilde{B}, W_i, Z_i)
	\! + \! \beta \NM{ Z_i }{1}\\
	& \text{s.t.} & 
	\nonumber 
	\NM{\CC{\iFFT{\tilde{B}(:,r)}}}{2} \! \le \! 1, \;r =1,\dots,R,
	\\
	&& 
	W_i \in \mathcal{W}, \; i=1,\dots,N, \nonumber 
	\end{eqnarray}
	where 
	$\tilde{f}_i(\tilde{B}, W_i, Z_i)
	\equiv 
	\frac{1}{2 P}
	\NM{\tilde{x}_i 
		- 
		\sum_{r=1}^{R}
		\tilde{B}(:,r)
		\odot 
\tilde{Y}_i(:,r) }{2}^2$, and
$\tilde{Y}_i(:,r) \equiv \FFT{Z_iW_i^\top(:,r)}$.
\end{proposition}
The spatial-domain base filters can be recovered 
from $\tilde{B}$ 
as $B(:,r) = \mathcal{C}(\iFFT{\tilde{B}(:,r)})$.

\subsubsection{Obtaining $\tilde{B}_t$}
\label{sec:bt}

From (\ref{eq:cscsd}),
$\tilde{B}_t$ can be obtained by solving the subproblem:
\begin{eqnarray*}
& 
\min_{\tilde{B},\bar{V}}& 
\frac{1}{2tP}\sum_{i=1}^{t} 
\NML{\tilde{x}_i - \sum_{r=1}^{R}\tilde{B}(:,r)\odot\tilde{Y}_i(:,r)}{2}^2
\label{eq:cscsd_base_fre} 
\\
& \text{s.t.} &
\FFT{\bar{V}(:,r)} = \tilde{B}(:,r), 
r =1,\dots,R,
\nonumber \\
&&\NM{\CC{\bar{V}(:,r)}}{2} \le 1, 
r =1,\dots,R,
\notag
\end{eqnarray*}
where $\bar{V}$ is an auxiliary variable. 
This is of the same form as \eqref{eq:csc_dic}. Hence,
analogous to \eqref{eq:ocsc_dic},
$\tilde{B}_t$ can be obtained as:
\begin{eqnarray} 
&\min_{\tilde{B},\bar{V}} &
\frac{1}{2tP}\sum_{i=1}^{t}
\sum_{p=1}^{P}
\tilde{B}(p,:)\bar{H}_t(:,:,p) \tilde{B}^{\dagger}(:,p)
\notag 
\\
&&
- 2\tilde{B}(p,:)\bar{G}_t(:,p)
\label{eq:cscsd_filter}
\\
& \text{s.t.} &
\FFT{\bar{V}(:,r)} = \tilde{B}(:,r), 
r =1,\dots,R,
\nonumber \\
&&
\NM{\CC{\bar{V}(:,r)}}{2} \le 1, 
r =1,\dots,R,
\notag
\end{eqnarray}
where
$\bar{H}_t(:,:,p)
\!=\!\!\frac{1}{t}\sum_{i=1}^{t}\tilde{Y}_i^{\top}(:,p)\tilde{Y}^\star_i(p,:)\in\mathbb{R}^{R\times
	R}$, and
$\bar{G}_t(:,p)
=\frac{1}{t}\sum_{i=1}^{t}\tilde{x}^\star_i(p)\tilde{Y}_{i}^{\top}(:,p)\in\mathbb{R}^{R}$.
They can be incrementally updated as
\begin{align}
\!\!\!\!
\bar{H}_t(:,:,p)
& = \frac{t - 1}{t} \bar{H}_{t-1}(:,:,p)
\! + \!
\frac{1}{t}\tilde{Y}_t^{\top}(:,p)\tilde{Y}^\star_t(p,:),
\label{eq:h1_update_cscsd}
\\
\!\!\!\!
\bar{G}_t(:,p)
& = \frac{t - 1}{t} \bar{G}_{t-1}(:,p)+\frac{1}{t}\tilde{x}^\star_t(p)\tilde{Y}_{t}^{\top}(:,p). \label{eq:h2_update_cscsd}
\end{align}
Problem~(\ref{eq:cscsd_filter})
can then be solved using ADMM as in \eqref{eq:ocsc_dic}.

\begin{table*}[ht]
\vspace{-10px}
\caption{Comparing the proposed SCSC algorithm with other scalable CSC algorithms on per iteration cost. 
For CCSC, 
the cost is measured per machine, and $L$ is the number of machines in the distributed system.
Usually, $N\gg K\gg R$, and $P\gg M$. 
}
\centering
\begin{tabular}{c|c|c|c}
	\hline
	                        $\!\!$                          &         $\!\!$space $\!\!$          &               $\!\!$code update time$\!\!$                &                   $\!\!$ filter update time $\!\!$                   \\ \hline
	             OCSC 
	\cite{wang2018online}               &              $O(K^2P)$              &                     $O(KP+KP\log P)$                      &                          $O(K^2P+KP\log P)$                          \\ \hline
	$\!\!\!$OCDL-Degraux 
	\cite{degraux2017online}$\!\!\!$ &           $O(K^2M^2+KPM)$           &                        $O(K^2P^3)$                        &                   $\!\!$$O(K^2PM^2+KPM)$$\!\!\!\!$                   \\ \hline
	            OCDL-Liu 
	\cite{liu2017online}             &              $O(K^2P)$              &                     $O(KP+KP\log P)$                      &                          $O(K^2P+KP\log P)$                          \\ \hline
	         CCSC 
	\cite{choudhury2017consensus}           & $\!\!$ $O(\frac{NKP}{L}+KP)$ $\!\!$ & $\!\!$ $O(\frac{NKP}{L}+NKP\log(\frac{P}{L}))$ $\!\!\!\!$ & $\!\!$$O(\frac{NK^2P}{L}+\frac{NKP}{L}\log(\frac{P}{L}))$ $\!\!\!\!$ \\ \hline
	                         SCSC                          &              $O(R^2P)$              &                     $O(RKP+RP\log P)$                     &                          $O(R^2P+RP\log P)$                          \\ \hline
\end{tabular}
\label{tab:cost}
\end{table*}

\subsubsection{Obtaining $W_t$ and $Z_t$}
\label{sec:base_code}


With the arrival of $x_t$, 
we fix
the base filters to $\tilde{B}_{t-1}$ learned at the last iteration, and
obtain 
$(W_t,Z_t)$ 
from \eqref{eq:cscsd} 
as:
\begin{align}
\!\!\!
\min_{W, Z} F(W, Z) 
\! \equiv \! \tilde{f}_t( \tilde{B}_{t-1}, W, Z ) 
\! + \! I_{ \mathcal{W} }( W )
\! + \! \beta \NM{Z}{1},
\label{eq:cscsd_code}
\end{align}
where $I_{\mathcal{W}}(W)$ is the indicator function on $\mathcal{W}$
(i.e., $I_{\mathcal{W}}(W) = 0$ if $W\in \mathcal{W}$ and $\infty$ otherwise).

As in the CSC literature, it can be shown that
ADMM can also be used to solve
(\ref{eq:cscsd_code}).
While 
CSC's
code update
subproblem in
\eqref{eq:csc_code} 
is convex, problem
(\ref{eq:cscsd_code}) 
is nonconvex and 
existing convergence results for ADMM 
\cite{wang2015global}
do not apply.

In this paper, we will instead use the
nonconvex and inexact accelerated proximal gradient (niAPG) algorithm \cite{yao2017efficient}.
This is
a recent proximal algorithm for nonconvex problems.
As the regularizers on $W$ and $Z$ 
in \eqref{eq:cscsd_code} are
independent,
the proximal step
w.r.t. the two blocks
can be performed
separately 
as:
$( \Px{ I_{ \mathcal{W} }( \cdot ) }{ W },\Px{\beta \NM{\cdot}{1}}{Z} )$
\cite{parikh2014proximal}.
As shown in 
\cite{parikh2014proximal}, these individual proximal steps
can be easily computed 
(for $\mathcal{W} = \mathcal{W}_{\ell 1}$ or $\mathcal{W}_{\ell 2}$).

\subsubsection{Complete Algorithm}
\label{sec:complexity}

The whole procedure,
which will be called ``Sample-dependent Convolutional Sparse Coding (SCSC)", 
is shown in Algorithm~\ref{alg:cscsd}. 
Its space complexity, which is
dominated by $\bar{H}_t$ and $\bar{G}_t$,
is $O(R^2P)$. 
Its per-iteration time complexity is
$O(RKP+RP\log P)$, where
the $O(RKP)$ term is due to gradient computation, and
$O(RP\log P)$ 
is due to FFT/inverse FFT.
Table~\ref{tab:cost} compares its
complexities with those of
the other online 
and distributed CSC algorithms.
As can be seen, 
SCSC 
has much lower time and space
complexities
as $R\ll K$.

\begin{algorithm}[ht]
	\caption{Sample-dependent CSC (SCSC).}
	\begin{algorithmic}[1]
		\STATE {\bf Initialize} $W_0\in \mathcal{W}$, $B_0\in \mathcal{B}$, 
		$\bar{H}_0 = \mathbf{0}$,
		$\bar{G}_0 = \mathbf{0}$;
		\FOR{$t = 1,2,\dots, T$}
		\STATE draw $x_t$ from $\{x_i\}$;
		\STATE $\tilde{x}_t = \FFT{x_t}$;
		\STATE obtain $W_t, Z_t$ using niAPG;
		\FOR{$r = 1, 2,\dots, R$}
		\STATE $\tilde{Y}_t(:,r) = \FFT{Z_tW_t^\top(:,r)}$;
		\ENDFOR
		\STATE update $\{\bar{H}_t(:,:,1),\dots,\bar{H}_t(:,:,P)\}$ using \eqref{eq:h1_update_cscsd};
		\STATE update $\{\bar{G}_t(:,1),\dots,\bar{G}_t(:,P)\}$ using \eqref{eq:h2_update_cscsd};
		\STATE update $\tilde{B}_t$ by \eqref{eq:cscsd_filter} using ADMM;
		\ENDFOR
  	    \FOR{$r = 1, 2,\dots, R$}
		\STATE
		$B_T(:,r) = \mathcal{C}(\iFFT{\tilde{B}_T(:,r)}) $;
		\ENDFOR
		\OUTPUT $B_T$.
	\end{algorithmic}
	\label{alg:cscsd}
\end{algorithm}



\section{Experiments}
\label{sec:expt}

Experiments are performed on 
a number of 
data sets (Table~\ref{tab:data_stat}). 
\textit{Fruit} and \textit{City} are two small
image data sets
that have been commonly used in the CSC literature \cite{zeiler2010deconvolutional,bristow2013fast,heide2015fast,papyan2017convolutional}.  
We use the default training and testing splits provided in \cite{bristow2013fast}.
The images are pre-processed as in 
\cite{zeiler2010deconvolutional,heide2015fast,wang2018online}, which includes
conversion to grayscale, feature standardization, local contrast normalization and edge tapering. 
These two data sets are small. 
In some experiments,
we will also 
use two larger data sets,
\textit{CIFAR-10} 
\cite{krizhevsky2009learning}
and 
\textit{Flower} 
\cite{nilsback2008automated}.
Following \cite{heide2015fast,choudhury2017consensus,papyan2017convolutional,wang2018online}, we set the filter size $M$ as $11\times11$, and the regularization parameter $\beta$ 
in (\ref{eq:csc})
as 1.

\begin{table}[ht]
    \vspace{-5px}
	\caption{Summary of the image data sets used.}
	\centering
	\begin{tabular}{c|ccc }
		\hline
		& size & \#training & \#testing\\ \hline
		\it Fruit &  100$\times$100  &     10     &     4     \\ \hline
		\it City  &  100$\times$100  &     10     &     4     \\ \hline
		\it CIFAR-10 &  32$\times$32  &   50,000    &   10,000   \\ \hline
		\it Flower  & 500$\times$500 &    2,040    &   6,149    \\ \hline
	\end{tabular}
	\label{tab:data_stat}
\end{table}


To evaluate efficacy of the learned dictionary,
we will mainly consider the task of image reconstruction as in
\cite{aharon2006rm,heide2015fast,sironi2015learning}. 
The reconstructed image quality is evaluated by
the testing peak signal-to-noise ratio 
\cite{papyan2017convolutional}:
$\text{PSNR}=\frac{1}{|\Omega|}\sum_{x_j \in \Omega }20\log_{10}\left( \frac{\sqrt{P}}{
	\NM{\hat{x}_j-{x}_j}{2} } \right)$,
where $\hat{x}_j$ is the reconstruction of $x_j$ from test set $\Omega$.
The experiment 
is repeated 
five times with different
dictionary initializations.

\subsection{Choice of $\mathcal{W}: \mathcal{W}_{\ell 1}$ versus $\mathcal{W}_{\ell 2}$}
\label{sec:choicel12}

First, 
we study the choice of $\mathcal{W}$ in Proposition~\ref{pr:constraint}.
We compare 
SCSC-L1, which uses	$\mathcal{W} =\mathcal{W}_{\ell 1}$,
with 
SCSC-L2,
which uses $\mathcal{W} = \mathcal{W}_{\ell 2}$.
Experiments are performed on \textit{Fruit} and \textit{City}.
As in \cite{heide2015fast,papyan2017convolutional,wang2018online}, the number of filters
$K$ is set to 100.
Recall the space complexity results in Table~\ref{tab:cost},
we define the compression ratio of SCSC relative to OCSC (using the same $K$) as $\text{CR} = (K/R)^2$.
We vary 
$R$ in $\{ K/20, K/10, K/9, \dots, K/2, K\}$.
The
corresponding 
CR is $\{ 400,100,81, \dots, 1 \}$.

\begin{figure}[ht]
\centering
\subfigure[\textit{Fruit}.]
{\includegraphics[width=0.2375\textwidth]{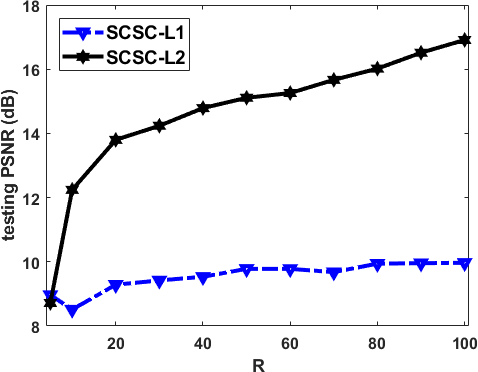}}
\subfigure[\textit{City}.]
{\includegraphics[width=0.2375\textwidth]{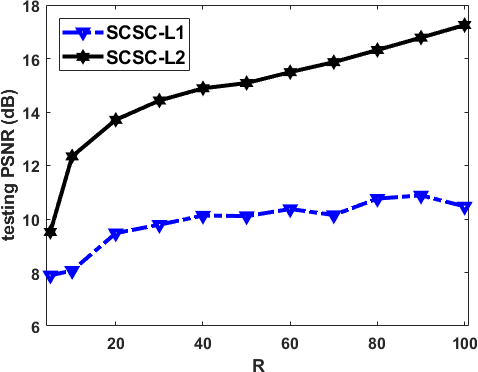}}

\vspace{-10px}

\caption{Testing PSNR's of SCSC-L1 and SCSC-L2 at different $R$'s on the \textit{Fruit} and \textit{City} data sets.}
\label{fig:cscsdl12}
\end{figure}

\begin{figure}[ht]
\centering

\subfigure[SCSC-L1.
\label{fig:cscsd_l1}]{\includegraphics[width=0.235\textwidth]{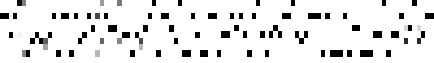} \label{fig:L1W}}
\subfigure[SCSC-L2.  \label{fig:cscsd_l2}]
{\includegraphics[width=0.235\textwidth]{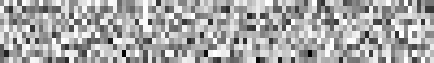}\label{fig:L2W}}
\vspace{3px}	
\includegraphics[width = 0.4\textwidth]{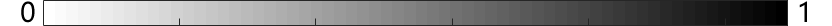}

\vspace{-10px}
\caption{Weight matrices obtained on a test sample from \textit{City}. Each column
corresponds to an original filter.
}
\label{fig:W}
\end{figure}

Results are shown in Figure~\ref{fig:cscsdl12}.
As can be seen,
SCSC-L1 is much inferior. 
Figure~\ref{fig:L1W}
shows
the weight $W_j$ obtained with $K=100$ and $R=10$ by SCSC-L1 
on a test sample $x_j$ from \textit{City}
(results on the other data sets are similar).
As can be seen, most of its entries are zero because of the sparsity induced by the $\ell_1$ norm.
The expressive power is severely limited
as typically only one base filter is used to approximate the original filter.
On the other hand,
the $W_j$ learned by SCSC-L2 is dense and has more nonzero entries
(Figure~\ref{fig:L2W}).
In the sequel, we will only focus  on
SCSC-L2, which will be simply denoted as
SCSC.

\subsection{Sample-Dependent Dictionary}
\label{sec:expts_sample_dependent}

In this experiment,
we set $K=100$, and compare SCSC with the following algorithms that use
sample-independent dictionaries:
(i) SCSC (shared): This is a SCSC  variant in which 
all $W_i$'s in \eqref{eq:x_two_level}
are the same.
Its optimization
is based on	alternating minimization.
(ii) Separable filters learned by tensor decomposition
(SEP-TD) 
\cite{sironi2015learning}, 
which is based on
post-processing
the  (shared)
dictionary learned by OCSC
as reviewed in Section~\ref{sec:post};
(iii) OCSC
\cite{wang2018online}:
the state-of-the-art online CSC algorithm.

\begin{figure}[ht]
\centering
\subfigure[\textit{Fruit}.  \label{fig:r_fruit}]
{\includegraphics[width=0.235\textwidth]{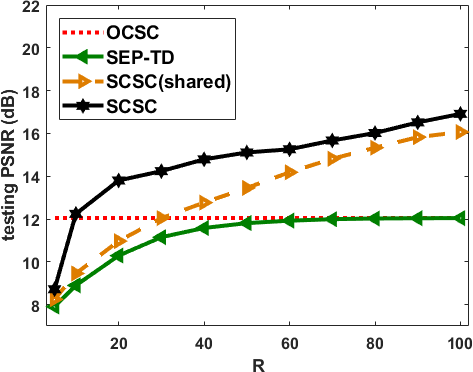}}
\subfigure[\textit{City}.  \label{fig:r_city}]
{\includegraphics[width=0.235\textwidth]{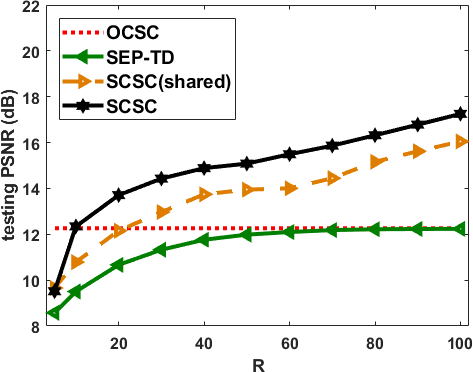}}

\vspace{-10px}
\caption{Testing PSNR vs $R$ for CSC algorithms using shared and sample-dependent
dictionaries.}
\label{fig:r_small}
\end{figure}

Results are shown in Figure~\ref{fig:r_small}.
As can be seen, 
SCSC always outperforms 
SCSC(shared) and
SEP-TD,
and outperforms OCSC 
when
$R=10$ 
(corresponding to $\text{CR}=100$) or above.
This demonstrates 
the advantage of using a sample-dependent dictionary.

\begin{figure}[ht]
\centering
\subfigure[\textit{Fruit}.  \label{fig:finetune_fruit}]
{\includegraphics[width=0.235\textwidth]{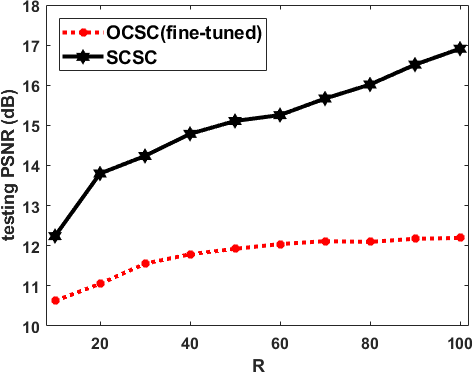}}
\subfigure[\textit{City}.  \label{fig:finetune_city}]
{\includegraphics[width=0.235\textwidth]{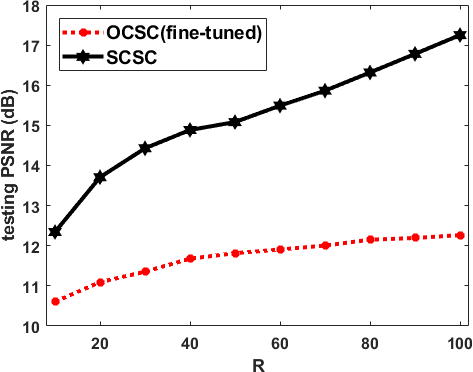}}

\vspace{-10px}
\caption{Comparison of SCSC and OCSC with fine-tuning.}
\label{fig:finetune_small}
\end{figure}

Next,
we compare against
OCSC
with 
fine-tuned filters,
which 
are also sample-dependent.
Specifically,
given test sample $x_j$, we first obtain its code $Z_j$ from
\eqref{eq:csc_code} with the learned dictionary $D$, and
then fine-tune $D$ by solving \eqref{eq:csc_dic} using the newly computed $Z_j$.
As in \cite{donahue2014decaf},
this is repeated for a few 
iterations.\footnote{In the experiments, we 
stop after five iterations.}
We set 
OCSC's $K$ 
to be equal to 
SCSC's
$R$,
so that the two methods take the same space (Table~\ref{tab:cost}).
The $K$ used in SCSC is still 100.
Results are shown in Figure~\ref{fig:finetune_small}.
As can be seen, though 
fine-tuning improves
the performance of OCSC slightly,
this approach of generating sample-dependent filters is still much worse than SCSC.

\subsection{Learning with More Filters}
\label{sec:k} 

Recall that SCSC allows the use of 
more filters (i.e., a larger $K$) because of its lower time and space complexities.
In this Section,
we demonstrate that this can lead to better performance.
We compare SCSC with 
two most recent batch and online CSC methods, namely,
	slice-based CSC (SBCSC) 
\cite{papyan2017convolutional} and
OCSC.
For SCSC, 
we set $R=10$ for \textit{Fruit} and \textit{City}, and $R=30$ for 
\textit{CIFAR-10}
and 
\textit{Flower}.

\begin{figure}[ht]
\centering

\subfigure[\textit{Fruit}.\label{fig:fruit10_more}]
{\includegraphics[width=0.235\textwidth]{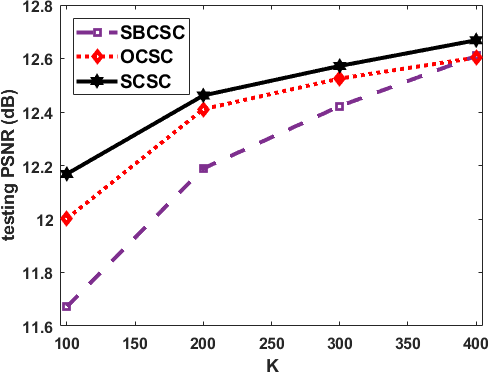}}
\subfigure[\textit{City}.  \label{fig:city10_more}]
{\includegraphics[width=0.235\textwidth]{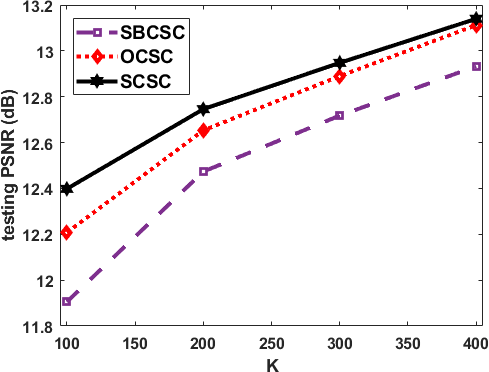}}

\vspace{-10px}

\subfigure[\textit{CIFAR-10}.\label{fig:cifar10_more}]
{\includegraphics[width=0.235\textwidth]{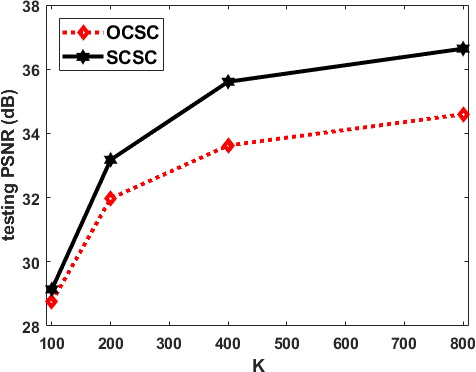}}
\subfigure[\textit{Flower}.  \label{fig:flower_more}]
{\includegraphics[width=0.235\textwidth]{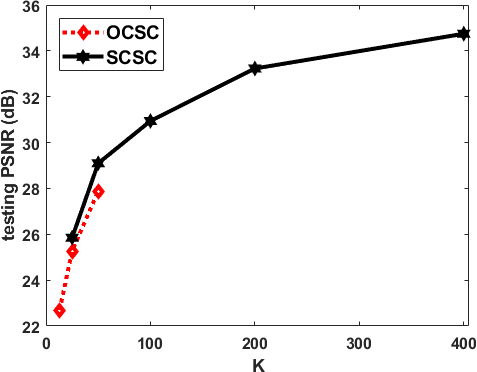}}		

\vspace{-10px}
\caption{Effect of $K$ on the testing PSNR. Note that SBCSC cannot be run on 
\textit{CIFAR-10} and \textit{Flower},
which are large.
For OCSC, it can only run up to $K=50$ on
\textit{Flower}.}
\label{fig:impactK_large}
\end{figure}

Figure~\ref{fig:impactK_large} shows the 
testing PSNR's at different 
$K$'s. 
As can be seen, a larger $K$ consistently leads to better performance for all 
methods.
SCSC allows the use of a larger $K$ because  of its much smaller memory
footprint.
For example, on \textit{CIFAR-10}, 
$\text{CR}=1024$
at $K=800$;
on \textit{Flower}, 
$\text{CR}=1600$
at $K=400$.

\subsection{Comparison with the State-of-the-Art}
\label{sec:expts_soa}

First, we perform experiments on the two smaller data sets of
\textit{Fruit} and \textit{City},
with $K=100$.
We set $R = 10$ (i.e., $\text{CR}=100$)
for SCSC. This
is compared with
the batch CSC algorithms, including
(i) deconvolution network (DeconvNet)
\cite{zeiler2010deconvolutional},
(ii) fast CSC (FCSC) \cite{bristow2013fast},
(iii) fast and flexible CSC (FFCSC) \cite{heide2015fast},
(iv) convolutional basis pursuit denoising (CBPDN) \cite{wohlberg2016efficient},
(v) the CONSENSUS algorithm \cite{sorel2016fast}, and
(vi) slice-based CSC (SBCSC) \cite{papyan2017convolutional}.
We also compare with the
online CSC algorithms, including
(vii) OCSC \cite{wang2018online},
(viii) OCDL-Degraux \cite{degraux2017online},
and
(ix) OCDL-Liu \cite{liu2017online}. 

\begin{figure}[ht]
\centering

\subfigure[\textit{Fruit}.  \label{fig:test_fruit}]
{\includegraphics[width=0.48\columnwidth]{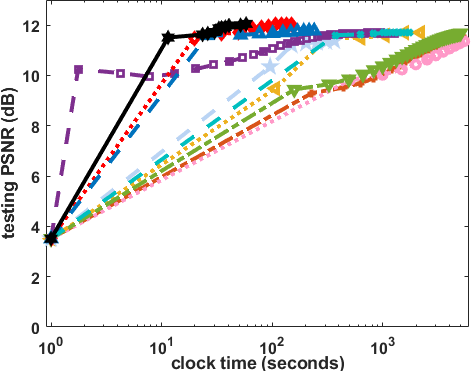}}
\subfigure[\textit{City}.  \label{fig:test_city}]
{\includegraphics[width=0.48\columnwidth]{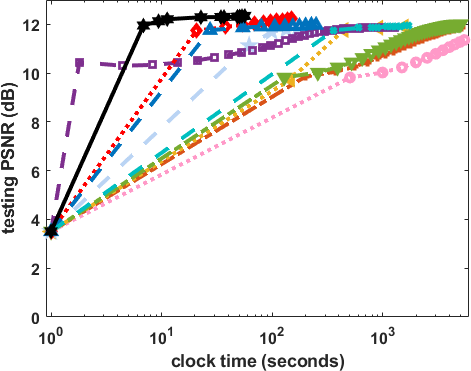}}

\includegraphics[width=0.8\columnwidth]{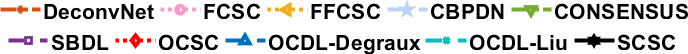}	

\vspace{-5px}
\caption{Testing PSNR on the small data sets. }
\label{fig:eval_small}

\end{figure}

Figure~\ref{fig:eval_small} shows convergence of the testing PSNR with clock time.
As also demonstrated in \cite{degraux2017online,liu2017online,wang2018online}, online CSC methods 
converge faster and have better PSNR 
than batch CSC methods.
Among the online methods, SCSC
has comparable PSNR as OCSC, but is faster and requires much less storage
($\text{CR}=100$).

\begin{figure}[ht]
\centering
\subfigure[\textit{CIFAR-10}.]
{\includegraphics[width=0.48\columnwidth]{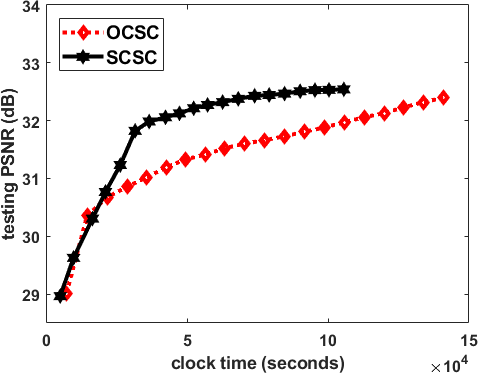}
\label{fig:testK_cifar10}}
\subfigure[\textit{Flower}.]
{\includegraphics[width=0.48\columnwidth]{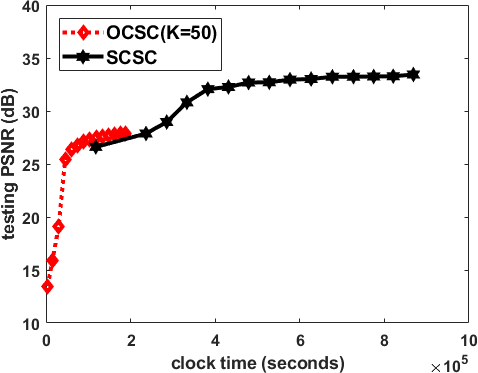}
\label{fig:testK_flower}}

\vspace{-10px}
\caption{Testing PSNR on the large data sets.}
\label{fig:large}
\end{figure}

Next, we perform experiments on 
the two large data sets,
\textit{CIFAR-10} and
\textit{Flower}.
All the batch CSC algorithms and two online CSC algorithms,
OCDL-Degraux and OCDL-Liu,
cannot handle such large data sets. Hence, we will only compare SCSC with OCSC.
On \textit{CIFAR-10},
we set $K=300$, and the corresponding $\text{CR}$ for SCSC is 100.
On \textit{Flower},
$K$ is still 300
for SCSC. However,
OCSC can only 
use $K=50$ 
because of its much larger memory footprint.
Figure~\ref{fig:large} shows convergence of the testing PSNR.  In both cases, SCSC significantly outperforms OCSC.  

\subsection{Higher-Dimensional Data}

In this section, we perform experiments on
data sets with 
dimensionalities larger than two.
To alleviate the large memory
problem,
Choudhury \etal \yrcite{choudhury2017consensus}
proposed the use of distributed algorithms.
Here, we 
show that SCSC can effectively handle these data sets using one single machine.

Experiments are performed on three data sets (Table~\ref{tab:data_stat_nd})
in \cite{choudhury2017consensus}.
The {\it Video} data set contains
image subsequences recorded in 
an airport
\cite{li2004statistical}.
The length of each video is 7,
and each image frame is of size $100\times 100$.  
The {\it Multispectral} data contains 
$60\times60$  
patches from
multispectral images 
(covering 31 wavelengths)
of real-world objects and materials
\cite{yasuma2010generalized}. 
The {\it Light field} data contains  
$60\times60$
patches of 
light field images 
on objects and scenes
\cite{kalantari2016learning}.
For each pixel,
the light rays are from $8\times 8$ different directions.
Following \cite{choudhury2017consensus}, we set the filter size $M$ to $11\times11\times 11$ for {\it Video}, $11\times11\times 31$ for {\it
Multispectral}, and $11\times11\times 8\times 8$ for {\it Light field}.

\begin{table}[ht]
\vspace{-5pt}
\caption{Summary of the higher-dimensional data sets used.}
\centering
\begin{tabular}{c|ccc}
	\hline
	&              size              & \#training & \#testing \\ \hline
	\it Video        &    100$\times$100$\times$7     &    573     &  143   \\ \hline
	\it Multispectral &     60$\times$60$\times$31     &    2,200    &  1,000  \\ \hline
	\it Light field & 60$\times$60$\times$8$\times$8 &    7,700    &  385   \\ \hline
\end{tabular}
\label{tab:data_stat_nd}
\end{table}

We compare SCSC with OCSC and the concensus CSC (CCSC) \cite{choudhury2017consensus} algorithms,
with $K=50$. 
For fair comparison, only one machine is used for all methods.
We do not compare with 
the batch methods and the two online methods 
(OCDL-Degraux and OCDL-Liu) as they 
are not scalable
(as already shown in Section~\ref{sec:expts_soa}).

Because of the small memory footprint of SCSC, we run it on a GTX 1080 Ti GPU in this experiment.
OCSC is also run on GPU for \textit{Video}. However, OCSC can only run on CPU for
		{\it Multispectral} and
{\it Light field}.
CCSC, which needs to access
all the samples and codes during processing,
can only be on CPU.\footnote{For \textit{Video}, the memory used (in GB) by CCSC,
OCSC, SCSC (with $R=5$) and SCSC (with $R=10$) are 28.73, 7.58, 2.66, and 2.87,
respectively. On {\it Multispectral}, they are 28.26, 11.09, 0.73 and 0.76; on {\it
Light field}, they are 29.79, 15.94, 7.26 and 8.88, respectively.}

Results are shown in Table~\ref{tab:highdim}.
Note that SCSC is the only method that can
handle the whole of {\it Video}, {\it Multispectral} and {\it Light field} data sets
on a single machine. In comparison,
CCSC can only handle a maximum of 
30 {\it Video} samples, 40 {\it Multispectral} samples, and 35 {\it Light field} samples.
OCSC can
handle the whole of {\it Video} and {\it Multispectral}, but
cannot converge 
in 2 days
when the whole \textit{Light field} data set is used.
Again, SCSC outperforms OCSC and CCSC.

\begin{table*}[ht]
\centering
\vspace{-10px}
\caption{Results on the higher-dimensional data sets. 
	PSNR is in dB and clock time is in hours. 
Timing results based on GPU
are marked with asterisks.
}
\begin{tabular}{c c | c c| c c|c c}
	\hline
	&   &\multicolumn{2}{c|}{\textit{Video}}& \multicolumn{2}{c|}{\textit{Multispectral}}&\multicolumn{2}{c}{\textit{Light field}}   \\
	\multicolumn{2}{c|}{}    & {\small PSNR}& time        & {\small PSNR} & time         & {\small PSNR} & time         \\ \hline
	\multicolumn{2}{c|}{CCSC}     & 20.43$\pm$0.11          & 11.91$\pm$0.07         & 17.67$\pm$0.14          & 27.88$\pm$0.07         &     13.70$\pm$0.09     & \textbf{8.99$\pm$0.11} \\ \hline
	\multicolumn{2}{c|}{OCSC}     & 33.17$\pm$0.01          & 1.41$\pm$0.04*          & 30.12$\pm$0.02          & 31.19$\pm$0.02         &   -           & -            \\ \hline
	\multirow{2}{*}{SCSC} & $R=5$& 35.30$\pm$0.02         & \textbf{0.73$\pm$0.02}* & 30.51$\pm$0.02          & \textbf{1.21$\pm$0.03}* &     29.30$\pm$0.03      & 11.12$\pm$0.07*         \\ \cline{2-8}
	&$R=10$& \textbf{38.02$\pm$0.03} & 0.81$\pm$0.01*          & \textbf{31.71$\pm$0.01} & 1.40$\pm$0.01*          & \textbf{31.70$\pm$0.02} & 17.97$\pm$0.05*        \\ \hline
\end{tabular}
\label{tab:highdim}
\end{table*}
As for speed, SCSC is the fastest. However, note that this is for reference only as SCSC is run on GPU while the others (except for OCSC on
{\it Video}) are run on CPU. Nevertheless, this still demonstrates an important advantage of
SCSC, namely that its small memory footprint can benefit from the use of GPU, while the others cannot.

\begin{table*}[hbt]
\centering
\vspace{-10px}
\caption{Testing PSNR (dB) on image denoising and inpainting.}
\label{tab:img_denoising_inpainting}
\begin{tabular}{c|c|c|c|c|c|c}
		\hline
		&\multicolumn{3}{c|}{denoising}&\multicolumn{3}{c}{inpainting}\\\hline
		& SBCSC & OCSC  &      SCSC & SBCSC & OCSC  &      SCSC     \\ \hline
		Wind Mill  & 14.88$\pm$0.03 & 16.20$\pm$0.03 & \textbf{17.27$\pm$0.02} &\textbf{29.76$\pm$0.13}  &29.40$\pm$0.14  &\textbf{29.76$\pm$0.08} \\ \hline
		Sea Rock   & 14.80$\pm$0.02 & 16.01$\pm$0.02 & \textbf{17.10$\pm$0.02} &24.92$\pm$0.06  &25.04$\pm$0.04  &\textbf{25.17$\pm$0.04}  \\ \hline
		Parthenon  & 14.97$\pm$0.02 & 16.33$\pm$0.01 & \textbf{17.44$\pm$0.03} &27.06$\pm$0.06  &26.79$\pm$0.04  &\textbf{28.04$\pm$0.04} \\ \hline
		Rolls Royce & 15.23$\pm$0.01 & 16.27$\pm$0.01 & \textbf{17.63$\pm$0.02} &24.96$\pm$0.13  &24.66$\pm$0.10  &\textbf{25.06$\pm$0.05} \\ \hline
		Fence    & 15.21$\pm$0.04 & 16.53$\pm$0.02 & \textbf{17.56$\pm$0.03} &26.81$\pm$0.05  &26.71$\pm$0.08  &\textbf{26.85$\pm$0.05} \\ \hline
		Car     & 16.90$\pm$0.01 & 18.05$\pm$0.03 & \textbf{20.06$\pm$0.05} &29.60$\pm$0.07  &29.40$\pm$0.09  &\textbf{30.44$\pm$0.04} \\ \hline
		Kid     & 14.90$\pm$0.01 & 16.21$\pm$0.02 & \textbf{17.22$\pm$0.03} &25.36$\pm$0.01  &25.42$\pm$0.07  &\textbf{25.67$\pm$0.07} \\ \hline
		Tower    & 14.89$\pm$0.02 & 16.19$\pm$0.01 & \textbf{18.36$\pm$0.05} &26.64$\pm$0.04  &26.48$\pm$0.06  &\textbf{26.96$\pm$0.03} \\ \hline
		Fish     & 16.40$\pm$0.01 & 17.40$\pm$0.01 & \textbf{18.61$\pm$0.02} &27.49$\pm$0.03  &26.98$\pm$0.08  &\textbf{27.23$\pm$0.07} \\ \hline
		Food     & 16.38$\pm$0.01 & 17.68$\pm$0.02 & \textbf{18.56$\pm$0.03} &29.96$\pm$0.05  &29.62$\pm$0.08  &\textbf{31.49$\pm$0.02} \\ \hline
	\end{tabular}
\vspace{-10px}
\end{table*}

\subsection{Image Denoising and Inpainting}
\label{sec:denoising_inpainting}

In previous experiments,
superiority of the learned dictionary is demonstrated by reconstruction of clean images. 
In this section,
we further examine the learned dictionary 
on two applications:
image denoising and inpainting. 
Ten test images provided by \cite{choudhury2017consensus} are used.
In denoising, we add Gaussian noise 
with zero mean and variance 0.01 to 
the test images (the average input PSNR is 10dB). 
In inpainting, we random sub-sample 50\% of the pixels as 0 (the average input PSNR is 9.12dB).  
Following \cite{heide2015fast,choudhury2017consensus,papyan2017convolutional}, 
we use a binary weight matrix to mask out positions of the missing pixels.
We use the 
filters learned from \textit{Fruit} in Section~\ref{sec:expts_soa}.
SCSC 
is compared
with 
(batch) 
SBCSC and 
(online)
OCSC.

Results are shown in Table~\ref{tab:img_denoising_inpainting}.
As can be seen,
the PSNRs obtained by SCSC are consistently higher than those by the other methods.
This shows that the dictionary, 
which yields high PSNR on image reconstruction,
also leads to better performance in other image processing applications.



\subsection{Solving \eqref{eq:cscsd_code}:
	niAPG vs ADMM}
\label{sec:niapg}

\begin{figure}[ht]
\centering
\includegraphics[width=0.58\columnwidth]{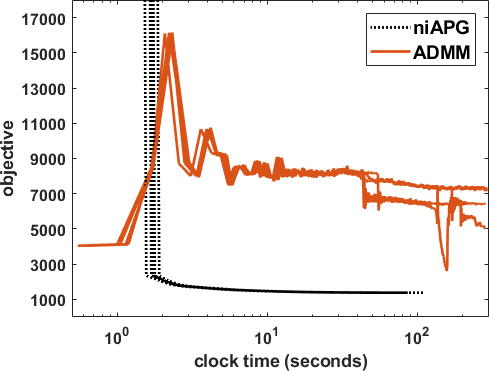}

\vspace{-10px}
\caption{Convergence of niAPG and ADMM on solving \eqref{eq:cscsd_code}.}
\label{fig:zw}
\vspace{-10px}
\end{figure}

\begin{figure}[ht]
\centering
\includegraphics[width=0.56\columnwidth]{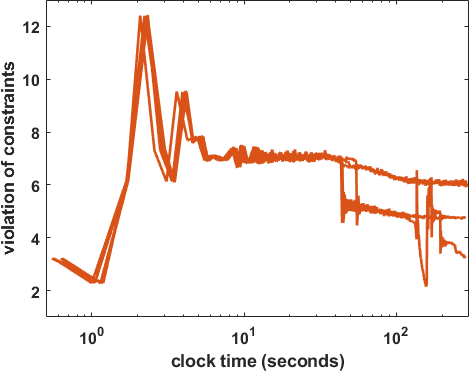}

\vspace{-10px}
\caption{Constraint violation in ADMM.}
\label{fig:zw_admm_residual}
\vspace{-20px}
\end{figure}

Finally, we compare the performance of ADMM 
and niAPG 
in solving subproblem~\eqref{eq:cscsd_code}. 
We use a training sample $x_i$ from \textit{City}.
The experiment is 
repeated 
five  times 
with different 
$(W_i,Z_i)$
initializations.
Figure~\ref{fig:zw} shows convergence of the objective in \eqref{eq:cscsd_code}
with time.
As can be seen, 
niAPG has fast convergence while
ADMM fails to converge.
Figure~\ref{fig:zw_admm_residual} shows $\NM{\tilde{Y}_i-\FFT{Z_iW^\top_i}}{F}^2$, which measures violation of the ADMM constraints,
with the number of iterations.
As can be seen, the violation does not go to zero, which 
indicates that ADMM does not converge.  

\section{Conclusion}

In this paper, we proposed a novel CSC extension, in which 
each sample has its own sample-dependent
dictionary constructed from 
a small set of 
shared base filters.
Using online learning,
the model can be  efficiently updated  with low time and space complexities.
Extensive experiments on a variety of data sets including large image data sets and higher-dimensional data sets all demonstrate its 
efficiency 
and
scalability.


\section*{Acknowledgements}

The second author especially thanks Weiwei Tu and Yuqiang Chen from 4Paradigm Inc. 
This research was supported in part by the Research Grants Council, Hong Kong, under Grant 614513,
and by the University of Macau Grant SRG2015-00050-FST.

\bibliographystyle{icml2018}
\bibliography{csc}

\cleardoublepage

\appendix
\section{Proofs}

\subsection{Proposition~\ref{pr:constraint}}
\label{app:constraint}
Since $B\in\mathcal{B}$, we have 
\begin{align}\label{eq:br}
\NM{B(:,r)}{2}\le 1,\forall r = 1,\dots, R,
\end{align}
so that
\begin{align}\label{eq:b}
\NM{B}{F}\le \sqrt{R}.
\end{align}

Therefore, when
{\setlist{leftmargin=6mm} \begin{itemize}
		\item[(i)]
		$W_i \in \mathcal{W}_{\ell 1}$:
		Use \eqref{eq:b}, we can write
		\begin{align*}
		\NM{BW_i(:,k)}{2}
		\le \NM{B}{F}\NM{W_i(:,k)}{2}
		\le \sqrt{R}\NM{W_i(:,k)}{2}.
		\end{align*}
		Thus,
		if $\sqrt{R}\NM{W_i(:,k)}{2}\le 1$, 
		then
		\begin{align*}
		\NM{BW_i(:,k)}{2}
		\le \sqrt{R}\NM{W_i(:,k)}{2}
		\le 1,
		\end{align*}
		which means $\NM{BW_i(:,k)}{2}\le 1$. 
		
		\item[(ii)]		
		$W_i \in \mathcal{W}_{\ell 2}$: 
		First,
		we have 
		$
		B W_i(:,k) = \sum_{r=1}^{R}W_i(r,k)B(:,r)$.
		Then,
		by Cauchy-Schwarz inequality,
		we have
		\begin{align}
		\NM{B W_i (:,k)}{2}
		& = \NML{\sum_{r=1}^{R}W_i(r,k)B(:,r)}{2}
		\notag
		\\
		& \le \sum_{r=1}^{R} \NM{ W_i(r,k) B(:,r) }{2} 
		\notag
		\\
		& \le \sum_{r=1}^{R}|W_i(r,k)| \NM{B(:,r)}{2} 
		\notag
		\\
		& = \sum_{r=1}^{R}|W_i(r,k)| = \NM{W_i(:,k)}{1},
		\label{eq:temp1}
		\end{align}
		where \eqref{eq:temp1} is due to \eqref{eq:br}.
		Therefore, if $\NM{W_i(:,k)}{1}\le 1$, $\NM{BW_i(:,k)}{2}\le 1$ holds.
\end{itemize}}

\subsection{Proposition~\ref{pr:cscsd}}
\label{app:cscsd}
Let $\tilde{B}(:,r) \equiv \FFT{B(:,r)}$, 
\eqref{eq:cscsd} is equivalent to \eqref{eq:cscsd_pre_app} since	
the following equations hold:
\begin{align}\notag 
& {f}_i (B,W_i,Z_i)
\\
&=
\frac{1}{2}
\NML{
	x_i 
	- 
	\sum_{k=1}^{K}
	\left( 
	\sum_{r=1}^{R}  W_i(r,k) {B}(:,r)
	\right)  *  Z_i(:,k)
}{2}^2,
\notag
\\\label{eq:ff_1}
&=
\frac{1}{2}
\NML
{x_i
	-
	\sum_{r=1}^{R}
	B(:,r)
	* 
	\left
	(\sum_{k=1}^{K}W_i(r,k)Z_i(:,k)
	\right
	)
}{2}^2,
\\\label{eq:ff_2}
&=
\frac{1}{2P}
\NML{\FFT{x_i}
	- 
	\sum_{r=1}^{R}\FFT{B(:,r)}
	\odot
	\FFT{Z_iW^\top_i(:,r)}}{2}^2
,
\\\notag
&=\tilde{f}_i(\tilde{B},W_i,Z_i),
\end{align}	
where \eqref{eq:ff_1} is due to 	
\begin{align*}
&\sum_{k=1}^{K}
\left(\sum_{r=1}^{R}W_t(r,k)B(:,r)\right)
* Z_t(:,k)
\\
=&
\sum_{r=1}^{R}
B(:,r)
* 
\left(\sum_{k=1}^{K}W_i(r,k)Z_i(:,k)\right).
\end{align*}
Then, 
\eqref{eq:ff_2} comes from the convolution theorem \cite{mallat1999wavelet}, i.e.,
\begin{align*}
\mathcal{F}({B}(:,r) * Z_iW_i^\top(:,r) )=\mathcal{F}({B}(:,r) )\odot\mathcal{F}( Z_iW_i^\top(:,r)),
\end{align*}
where ${B}(:,r)$ and $ Z_iW_i^\top(:,r)$ are first zero-padded to $P$-dimensional, 
and the Parseval's theorem \cite{mallat1999wavelet}: 
$\frac{1}{P} \NM{\FFT{x}}{2}^2 = \NM{x}{2}^2$ where $x\in\R^P$.

As for constraints, when $B$ is transformed to the frequency domain, it is padded
from $M$ dimensional to $P$ dimensional. Thus, we use $\CC{\iFFT{\tilde{B}}}$ to
crop the extra dimensions to get back the original support.

\end{document}